\pdfoutput=1

\documentclass[11pt]{article}

\usepackage[]{EACL2023}

\usepackage{times}
\usepackage{latexsym}

\usepackage{tabularx}
\usepackage{lipsum}
\usepackage{makecell}
\usepackage{hhline}
\usepackage{booktabs}
\usepackage{hyperref}
\usepackage{natbib}
\usepackage{graphicx}
\usepackage{subcaption}
\usepackage{multirow}
\usepackage{longtable}

\usepackage[T1]{fontenc}

\usepackage[utf8]{inputenc}

\usepackage{microtype}

\usepackage{inconsolata}

\title{``Define Your Terms'' : Enhancing  Efficient Offensive Speech Classification with Definition}

\author{Huy Nghiem \\
  University of Maryland, College Park \\
  \texttt{nghiemh@umd.edu} \And 
  Umang Gupta \and Fred Morstatter  \\ 
  USC Information Sciences Institute \\
  \texttt{umanggup@usc.edu, fred@isi.edu}
  }
  
\defcitealias{mathew2021hatexplain}{HateXplain}
\defcitealias{elsherief2021latent}{Implicit\_hate}
\defcitealias{hartvigsen2022toxigen}{ToxiGen}
\defcitealias{vidgen-etal-2020-detecting}{Covid}

\begin{document}
\maketitle
\begin{abstract}
The propagation of offensive content through social media channels has garnered  attention of the research community. Multiple works have proposed various semantically related yet subtle distinct categories of offensive speech. In this work, we explore meta-learning approaches to leverage the diversity of offensive speech corpora to enhance their reliable and efficient detection. We propose a joint embedding architecture that incorporates the input's label and definition for classification via Prototypical Network. Our model achieves at least  75\% of the maximal F1-score while using less than 10\% of the available training data across 4 datasets. Our experimental findings also provide a case study of training strategies valuable to combat resource scarcity.

\end{abstract}

\section{Introduction}
While a vital channel for the dissemination of crucial information, social media platforms have also become hotbeds for hateful, and harmful expressions. Such offensive speech not only detracts from the quality of discourse but also poses tangible threats to marginalized and vulnerable groups, escalating existing social tensions. Multiple studies have observed the psychological harms to marginalized communities perpetuated by offensive content in the digital space \citep{saha2019prevalence, stefuanițua2021hate}. However, the definition of offensive speech varies between contexts, and across publications that study this problem. A common challenge with offensive speech is the lack of a unifying definition, with conceptually related but definitively distinct categories proposed in literature: \textit{Hate, Abusive, Aggressive, Toxic, Offensive, Cyberbullying etc.} \citep{poletto2021resources, yin2021towards}. While earlier research focused on binary classification,  more current works have explored offensive categories in higher granularity and semantic diversity  \citep{mullah2021advances, caselli2021hatebert, mozafari2020bert, elsherief2021latent, yin2021towards}.

Though an active area of research in the Natural Language Processing (NLP) community, accurate and reliable detection of offensive speech often requires significant amount of training data \citep{vidgen2020directions, goodfellow2016deep}. For these tasks, the typical pipeline of data collection involves gathering a candidate corpus based on a set of relevant keywords, then soliciting task-specific labels for them via crowdsourcing or expert annotation \citep{vidgen2020directions, paullada2021data}. Demographics of annotators may be different from one dataset to the next, including platforms (e.g. Amazon Mechanic Turk, Prolific), payments, levels of education, languages and cultural backgrounds \citep{founta2018large}. 

As offensive content is frequently linked to real world events, there exists a need for appropriately tailored datasets. Nevertheless, constructing a sufficient amount of labelled data often proves a resource-intensive challenge \citep{poletto2021resources, founta2018large, toraman2022large}. On the other hand, there exists a plethora of available data on similar yet categorically distinct areas of offensive speech. We set out with the objective to discover suitable techniques capable of leveraging existing datasets to \textit{efficiently and reliably adapt} to new domains of offense content. 

To this end, we compile from literature a collection of 14 relevant datasets, which allow us to perform a battery of testing on various pre-training and meta-learning approaches to assess their efficacy and robustness in classification of offensive content. We also experiment with different model architectures to incorporate label information to enhance knowledge transference at multiple levels of data availability. We introduce \textbf{JE\_ProtoNet}, a \textit{joint embedding} based on Prototypical Network which utilizes \textbf{definition of label categories} and exhibit competitive performance across 4 test sets while using a fraction of the available training data. To the best of our knowledge, this work is the first in literature that harnesses label definition in offensive speech detection. Our experiments provide a case study on the trade-off between sample efficiency and performance, with findings potentially applicable to any classification task where categories entail more nuanced expression beyond simple labels. Based on empirical findings, we provide a set of recommendations to leverage our approach to enhance the efficiency of offensive speech classification.

\section{Related Work}
\subsection{Annotation with Instructions}  High-quality annotation is crucial to the development of offensive speech classifiers. Annotators' implicit biases and disagreements could be propagated and even magnified by downstream models \citep{waseem2016you, vidgen2020directions, davani2023hate, akhtar2020modeling}. Explicitly priming annotators with clear instructions and definitions have been shown to reduce biases and enhance inter-annotator agreements 
\citep{sap2019risk, waseem2016you, parmar2023don}. 

\subsection{Cross-Dataset Transference}
The diversity of  datasets  on offensive speech has prompted researchers to investigate their generalizability. Models' performance tend to significantly drop when applied to out-of-domain dataset \citep{bansal2019proceedings, yin2021towards}. \citet{fortuna2021well}'s extensive study revealed that cross-dataset transference is highly influenced by their semantic similarity. Some works, such as HateBERT and fBERT, pre-trained the base model on specialized corpora to allow better adaptation to new datasets \citep{caselli2021hatebert, sarkar2021fbert}.

Model architecture and fine-tuning strategy could also enhance transferrability. \citet{mozafari2022cross} applied Model-Agnostic Meta-Learning (MAML) and Proto-MAML to BERT-based (\citet{devlin2018bert} ) models and observed improvements in few-show cross-lingual hate speech detection. \citet{kim-etal-2022-generalizable} used contrastive learning to enhance detection of implicit hate speech detection across three benchmarks. \citet{tran2020habertor} constructed HABETOR with fewer parameters but still demonstrated good generalizable performance across 2 out-of-domain datasets.  

\subsection{Label-Aware Classification}
The idea of constructing label embedding was pioneered by \citet{tang2015pte} in their work on Predictive Text Embedding. \citet{wang2018joint} followed up with Label-Embedding Attentive Model (LEAM), a joint embedding of words and labels downstream classification task. More recently, \citet{xiong2021fusing} leveraged BERT's self-attention mechanism for classification by concatenating labels' tokens directly into their respective inputs. \citet{luo2021don} took this idea further in their method Label-semantic Augmented Meta-Learner (LaSAML) via Prototypical Network, a framework capable of few-shot text classification. 

\section{Data Collection and Processing}
\subsection*{General Criteria}
Our goal is to leverage existing datasets to adapt to new domains of offensive content in a reliable and label-efficient manner.
To this end, we survey literature\footnote{https://hatespeechdata.com} to identify relevant existing datasets on offensive speech and related topics. We filter our options based on the following criteria: size (> 10,000 samples), diversity of label categories (or, the nature of offensive text these labels capture), availability of definitions and instructions, along with method of annotation. We strive to incorporate a sufficient number of categories related to offensive speech with distinct levels of granularity. When definitions of label categories are unavailable in the original work, we solicit this content from their authors. Detailed definitions for the labels are included in Tables~\ref{apx:train_def} and~\ref{apx:test_def} of the Appendix. Ultimately  14 datasets are chosen (Table~\ref{tab:datasources}), with the 4 below held out for final testing of the models and are not used for any pre-training.

\noindent \textbf{ToxiGen} is a large-scale machine-generated dataset by demonstration-based prompting. \citet{hartvigsen-etal-2022-toxigen} controlled machine generation to create a corpus of \textit{Benign} and \textit{Toxic} texts that cover 13 identity groups. In addition to its unique nature of construction, this dataset is included as a representative for binary classification tasks. \\
\textbf{HateXplain} is constructed by \citet{mathew2021hatexplain} with an emphasis on explanability. The authors asked annotators to highlight the span of tokens, called \textit{rationales}, that contribute to their selection of the labels. This dataset shares the same label space with \citet{davidson2017automated}, yet with different definitions for each term. \\
\textbf{Implicit\_hate} is developed by \citet{elsherief-etal-2021-latent} to fill the gap in the literature with respect to negative sentiments expressed in coded or indirect language. As the corresponding authors posited, detecting implicit hate speech is regarded as more challenging than its overt counterpart. \\
\textbf{Covid} focuses on the  rise of anti-Asian sentiments fueled by the COVID-19 pandemic \citep{vidgen-etal-2020-detecting}. Among other COVID-related hate speech corpora \citep{nghiem2021stop, he2021racism}, this dataset arguably considers the most nuanced categories of East Asian entities. 

\subsection*{Train Set Sampling}
The remaining 10 datasets are reserved for meta-training. Data "in the wild" tends to have considerably different distributions with very low representation of the offensive classes~\citep{poletto2021resources}. 
Further, the offensive content is frequently deleted from the platforms, making retrieval for research even more challenging \citep{poletto2021resources, vidgen2021learning}. Some datasets in our collections contain classes that suffer from extremely low prevalence. To alleviate these problems, we only select label classes that have clear definitions and significant samples relative to their respective set. Then, we employ stratified sampling to create a subset while maintaining as close to an equal distribution between classes as feasible. These sample sizes are reflected in Table \ref{tab:datasources}, with a final tally of 82,000. Finally, we perform pre-processing steps to standardize texts (details in Appendix \ref{apx:preprocess}).

\begin{table*}[t]
\footnotesize
\centering  
\renewcommand{\arraystretch}{1.35}  
\newcolumntype{C}[1]{>{\centering\arraybackslash}p{#1}}
\begin{tabular}{p{3.1cm}C{1.cm}C{1cm}p{1.5cm}C{1.25cm}p{5.7cm}}
\toprule  
\textbf{Dataset} & \textbf{Total Size} & \textbf{Sample Size} & \textbf{Platform} & \textbf{Annotation\ Method} & \textbf{Selected Labels} \\
\midrule  
\citealp{waseem2016hateful} & 16,914 & 3,000 & Twitter & E & Offensive, Not Offensive \\
\citealp{golbeck2017large} & 20,360 & 10,000 & Twitter & E & Harrassment, Not Harrassment \\
\citealp{davidson2017automated} & 24,800 & 5,000 & Twitter & C & Hate Speech, Offensive, Normal \\
\citealp{kumar2018aggression} & 15,000 & 10,000 & Facebook & - & Overly Aggressive, Covertly Aggressive, Non-Aggressive \\
\citealp{founta2018large} & 80,000 & 10,000 & Twitter & C & Normal, Abusive Language, Hate Speech \\
\citealp{zampieri2019predicting} & 14,100 & 6,000 & Twitter & C & Targeted Insult, Untargeted Insult, Not Offensive \\
\citealp{basile2019semeval} & 13,000 & 8,000 & Twitter & C & Hate Speech, Not Hate Speech \\
\citealp{sap2019social} & 44,671 & 10,000 & Reddit, Twitter, Gab, Stormfront & C & Offensive, Not Offensive \\
\citealp{vidgen2021learning} & 41,255 & 10,000 & - & C & Derogation, Animosity, Threatening, Support for Hateful Entities, Dehumanization, Neutral \\
\citealp{toraman2022large} & 100,000 & 10,000 & Twitter & E & Offensive, Hate, Normal \\ \hline
\citetalias{hartvigsen2022toxigen}& 274,186 & \textit{2,740} & Synthetic & - & Toxic, Benign \\
\citetalias{mathew2021hatexplain} & 20,148 & \textit{2,000} & Gab, Twitter & C & Hate Speech, Offensive, Normal \\
\citetalias{elsherief2021latent} & 6,346 & \textit{1,340} & Twitter & E, C & White Grievance, Incitement to Violence, Inferiority Language, Irony, Stereotypes and Misinformation, Threatening and Intimidation \\
\citetalias{vidgen-etal-2020-detecting} & 20,000 & \textit{2,000} & Twitter & E & Hostility against an East Asian Entity, Criticism of an East Asian Entity, Discussion of East Asian Prejudice, None of the Above \\ \hline
\textbf{Total} & 688,587 & 82,000 & & & \\
\bottomrule  
\end{tabular}
\caption{General information about the compiled data sources. For \textit{Annotation Method}, 
E stands for \textit{Expert}, where trained annotators are selected for labeling,  and C for \textit{Crowdsource}, a setting that employs a larger, typically non-specialized pool of workers. The last 4 datasets are reserved for eventual evaluation. The sample size of test sets (italicized) refers to values used for the final evaluation and is not included in the Total Size column.}
\label{tab:datasources}
\end{table*} 

\section{Experimental Setup}
\label{section4}
From here on, we refer to the datasets as \textit{domains}. With the goal of investigating the potential benefits of learning from semantically related but distinct data, our general experiment pipeline consists of first pretraining a model on the 10 reserved domains using different techniques \footnote{Our code repository is available at: \href{https://github.com/hnghiem-usc/define_your_terms}{https://github.com/hnghiem-usc/define\_your\_terms}}. Then, the model is fine-tuned and evaluated on \textit{each} of the 4 aforementioned test domains. More specifically, we hold out a portion of the test domains using the ratio described in their original publications (Table~\ref{tab:datasources}). We then perform K-shot sampling of the remaining data to fine-tune the pre-trained models where K $\in \left\{ 16, 32, 64, 128, 256 \right\}$. We used $K=64$ from the leftover data to select hyperparameters (details in Table \ref{apx:hyper} of the Appendix). Finally, the fine-tuned model is tested on the held-out dataset. The following sections describe our pretraining approaches.

\subsection{Baselines} 
We select base RoBERTa (Robustly Optimized BERT approach) as implemented by the Huggingface library to be the main structure of our model due to its strong performance on related sentiment classification tasks~\citep{liu2019roberta, elsherief2021latent, poletto2021resources}. Already pretrained on a large English corpus in an unsupervised fashion, this version of RoBERTa contains 12 layers of transformer blocks, 12 attention heads, and approximately 125 million trainable parameters. 

The baseline models\footnote{We use Huggingface's RobertaForSequenceClassification implementation} all use the [CLS] token from the embedding as input to the classification head -- a fully connected layer -- to produce logit scores for each label. The model seeks to minimize the Cross-Entropy loss, with parameters updated via \texttt{AdamW} Optimizer. The simplest baseline, \textbf{RoBERTa\_untrained} refers to training 
with only K samples from the test domains (K-shot learning), then evaluating on the held-out portion without using any form of pretraining. 

Inspired by \citet{gururangan2020don}, the next variant, \textbf{RoBERTa\_retrained}, trains the model on each of the test domain's entire (non-sampled) training set using the Mask Language Model's objective in a self-supervised manner, before being further fine-tuned through supervised learning with the K-shot samples. 

Finally, \textbf{RoBERTa\_binary}, incorporates the 82,000 samples in a simple fashion. We unify the different domains by collapsing the disparate label spaces into a binary mapping: all non-neutral categories into \textit{Offensive}, and the rest into \textit{Not Offensive}. The model is pretrained on this unified dataset using the supervised learning objective before being fine-tuned in a K-shot way on in-domain samples. Additionally, we also K-shot fine-tune then evaluate out-of-the-box \textbf{HateBERT} \citep{caselli2021hatebert} for comparison.

\subsection{Meta-Learning Settings}
In this section, we explore various meta-learning frameworks as a means of pre-training. The following frameworks all simulate N-way K-shot learning, where N is the number of classes (labels) in a domain, and K is the number of samples per class. Each learner (model) $f$ is parameterized by $\theta$, of which we seek to optimize over the classification tasks using the 10 reserved domains. 

At each training episode, a \textit{support} and \textit{query} set of the same size is sampled from a domain $\mathcal{D}_i$, where $i \in \left\{1,2, ..., 10\right\}$ for each of the reserved domains. For meta-training, K is restricted to $\left\{16, 32, 64, 128\right\}$ shots to accommodate domains with high number of categories. Since meta-training is computationally demanding, we train models on a single fixed seed and report aggregate results by K-shot fine-tuning the meta-trained model with 5 random seeds.

\subsubsection*{Training Without Label Information} 
In this standard setting, the learner's inputs do not incorporate any label information.  

\noindent\textbf{Prototypical Network}, or ProtoNet, is a metric-based meta-learning framework \cite{snell2017prototypical}. We use RoBERTa's [CLS] token as the encoded representation of each input. For each class $c \in \mathcal{C}$ in domain $\mathcal{D}_i$, a prototype $\mathbf{v}_c$ is constructed by taking the mean of all K samples:
\begin{equation}
    \mathbf{v}_c = \frac{1}{|\mathcal{S}_c|} * \sum_{(\mathbf{x}_i, y_i)\in \mathcal{S}_c} f_\theta(\mathbf{x}_i)
\end{equation}
where $\mathcal{S}_c$ denotes the support set for which $y_i = c$. Distribution over the classes is calculated by taking the softmax over the inverse distances $d_\varphi$ (Eucliean in our work) between inputs' embedding and the prototypes.
\begin{equation}
p(y=c|\textbf{x}) = \frac{exp(-d_\varphi(f_\theta(\mathbf{x}), \mathbf{v_c}))}{\sum_{c' \in \mathcal{C}}{exp(-d_\varphi(f_\theta(\mathbf{x}), \mathbf{v_{c'}}))}}
\end{equation}
Input $\mathbf{x}$ is assigned the label of the nearest prototype.


\textbf{ProtoMAML}, an optimization-based framework, extends Model-Agnostic Meta-Learning (MAML, \citet{finn2017model}), which aims to learn a good initialization of the learner's base parameters $\theta$ that can quickly adapt to new tasks with limited data. During meta-training, MAML optimizes the model virtually using the support set, then evaluates the gradients on the query set with respect to the original parameters. Designing the classification layer with MAML is challenging when tasks have different label spaces. To circumvent this problem, \citealp{triantafillou2019meta} proposed ProtoMAML, which incorporates Prototypical Network's strengths by reformulating the softmax over Euclidean distances as a linear layer with with softmax. By setting the weights of the linear layer to twice the prototypes, and the biases to the to the negative of the prototypes, we obtain a classification layer that would be compatible with any domain. We implement First-Order ProtoMAML in this work to avoid the computational cost of obtaining second-order derivatives as in the original MAML algorithm.

\begin{figure}[!ht]
\centering
\begin{subfigure}{\linewidth}
    \centering
    \includegraphics[width=\linewidth]{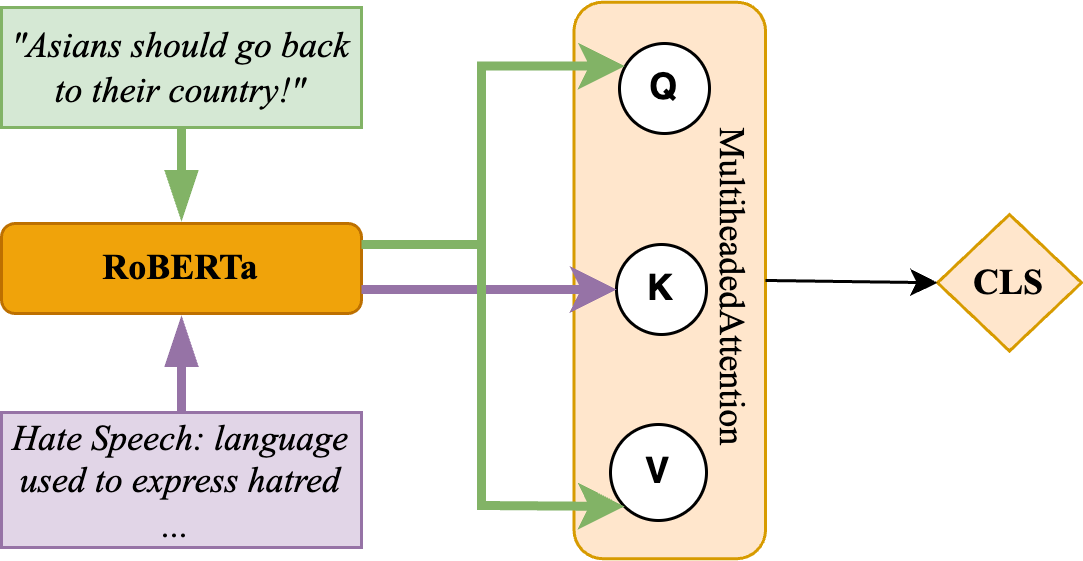}
\end{subfigure}
\caption{General architecture of JE\_ProtoNet. Hidden states of text input and its corresponding label and definition are obtained from RoBERTa, and then passed as Query, Value, and Key (color coded for traceability) to the Multiheaded Attention module.}
\label{fig:withlabel}
\end{figure}

\textbf{MLDG} (Meta-Learning for Domain Generalization) was proposed by \citet{li2018learning}. To learn a good initialization suited for generalization, $\mathcal{S}$ domains are split into disjoint sets $\overline{\mathcal{S}}$ and $\mathcal{S}^{'}$. During meta-training, 
MLDG updates the model's parameters virtually on tasks drawn from $\overline{\mathcal{S}}$ to using gradients $\nabla_\theta = \mathcal{F}^{'}_\theta(\overline{\mathcal{S}}, \theta)$. During meta-training, the model is virtually evaluated on tasks drawn from $\mathcal{S}^{'}$ to obtain loss $\mathcal{G}(\mathcal{S}^{'}; \theta^{'})$. The base model is optimized using both losses:
\begin{equation}
    \theta = \theta - \gamma\frac{\partial(\mathcal{F}(\overline{\mathcal{S}}, \theta) + \beta\mathcal{G}(\mathcal{S}^{'}; \theta - \alpha\nabla_\theta)}{\partial\theta})
\end{equation}
Inspired by \citet{ye2021train, kao2022maml}, the classification head takes the [CLS] token as input to pass through a linear layer of the same dimension (768), whose output is further connected to a final fully connected layer of shape (768,1). At the beginning of each training episode, this layer is duplicated accordingly to the required number of classes for each domain, with the parameters' weights set to 0. 

\subsection{Training with Label Information} 
In this setting, label information is directly incorporated into the training inputs in various configurations. ProtoNet is the sole chosen architecture because its metric-based nature limits overfitting on labels compared to other methods. Label incorporation only happens during meta-training and fine-tuning. During test time, no label information is available to the model. 

\noindent\textbf{ProtoNet\_Token} For each domain $\mathcal{D}_i$, we convert the label \textit{L\textsubscript{j}} into new token \textit{E\textsubscript{Lj}} for all \textit{j} in the label space. For labels that consist of multiple subwords, we construct \textit{E\textsubscript{Lj}} by averaging their token embeddings. Inspired by \citet{xiong-etal-2021-fusing} and \citet{si2020students}, we concatenate the token embedding \textit{E\textsubscript{T}} of input \textit{T}  with its corresponding label token \textit{E\textsubscript{Lj}}, separated by the [SEP] token. Labels from different domains but share identical textual representation would also share their token embeddings. 

\noindent\textbf{ProtoNet\_Label} In contrast, this setting concatenate the corresponding label \textit{L\textsubscript{j}} directly to the end of each text input \textit{T}, all of which are passed together to the model. This approach simplifies the label fusing process to create more discriminate representation of inputs \citep{luo2021don}. 

\noindent\textbf{ProtoNet\_Full} In this approach, we also utilize the definition associated with each label. Specifically, we construct the input to the model using the format [CLS] \textit{T} [SEP] \textit{L\textsubscript{j}} :  \textit{D\textsubscript{j}} [SEP], where \textit{D\textsubscript{j}} is the full definition of the corresponding label.

\noindent\textbf{JE\_ProtoNet} We construct an architecture that takes into consideration the compatibility between the text inputs and the labels' definitions via a joint embedding (illustrated in Figure \ref{fig:withlabel}). Each input \textit{T} is fed into the RoBERTa's backbone to obtain the hidden state representation \textit{H\textsubscript{T}}. Similarly, we obtain the hidden state \textit{H\textsubscript{D}} of the the corresponding label and definition sequence of the format \textit{L\textsubscript{j}} : \textit{D\textsubscript{j}} using the same model. We then pass \textit{H\textsubscript{T}} as the Query and Value input, and \textit{H\textsubscript{D}} as the Key into the attention module \footnote{We use Huggingface's MultiHeadAttention module} \citep{vaswani2017attention}, which consists of 3 attention heads. In contrast to the native self-attention mechanism seen in previous configurations, this setup allows the model to focus on certain aspects or parts of the input text semantically relevant to the given label definition.  Finally, we extract the [CLS] token from the output of joint embedding for downstream classification as in other ProtoNet settings. During testing, a blank string is passed in lieu of the definition.

\section{Results} 
\begin{figure*}[!ht]
\centering
\begin{subfigure}{\textwidth} 
    \centering
    \includegraphics[width=\linewidth]{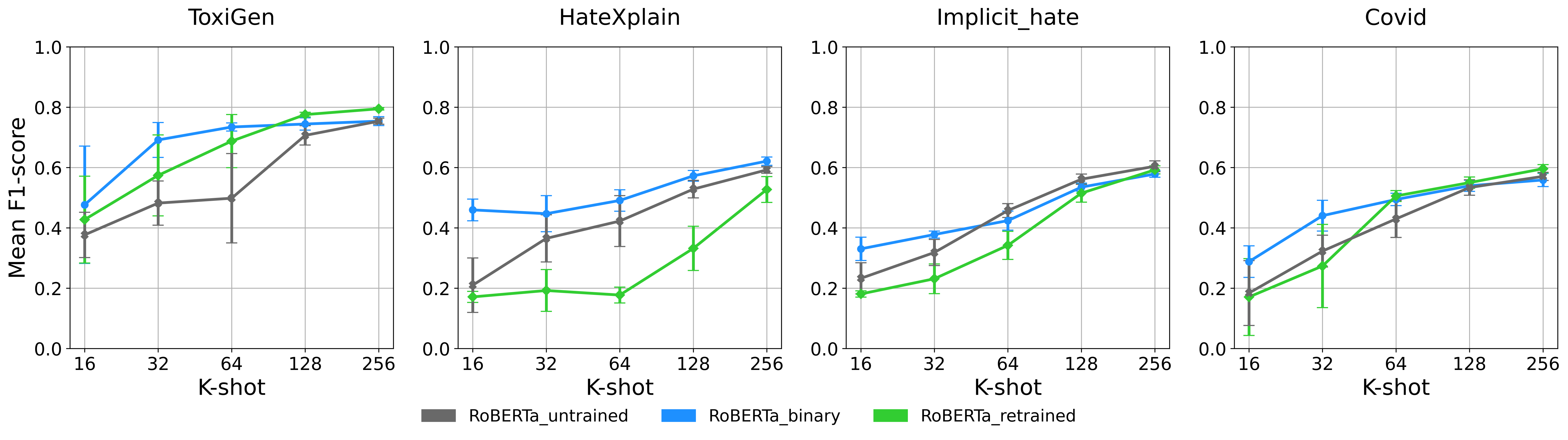}
    \caption{Results for Baseline RoBERTa models}
    \label{fig:resulta}
\end{subfigure}
\begin{subfigure}{\linewidth}
    \centering
    \includegraphics[width=\linewidth]{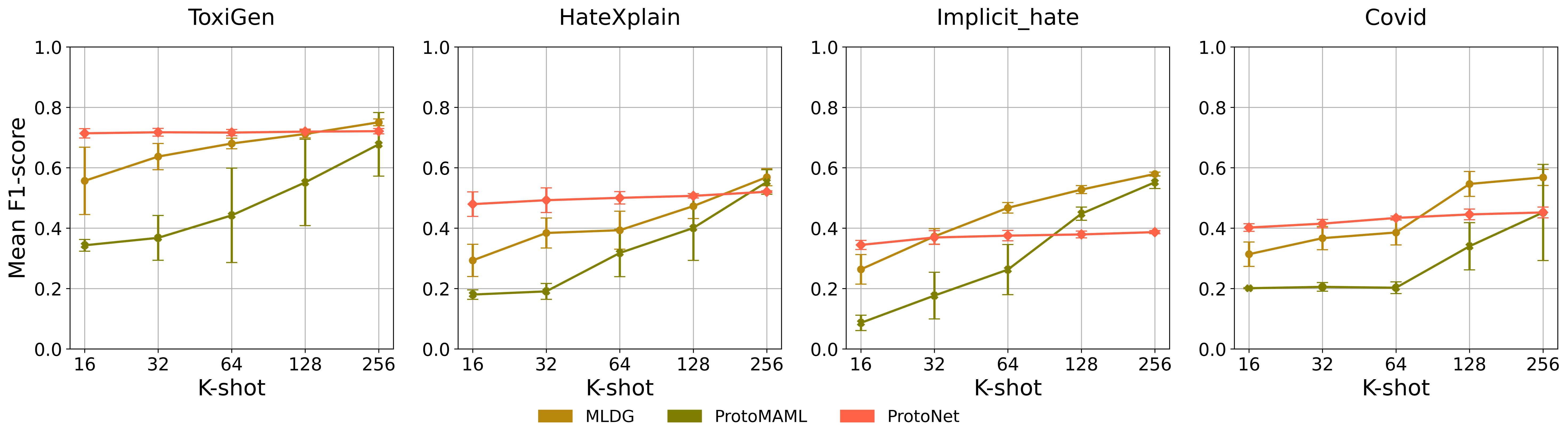}
    \caption{Results for Meta-Learning models without labels}
    \label{fig:resultb}
\end{subfigure}
\begin{subfigure}{\linewidth}
    \centering
    \includegraphics[width=\linewidth]{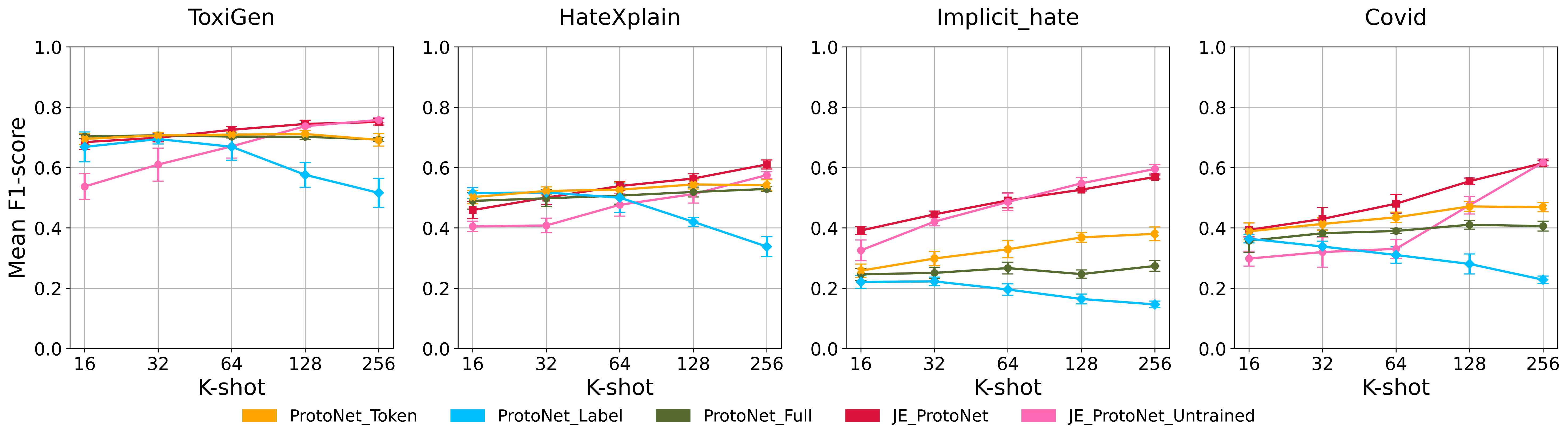}
    \caption{Results for ProtoNet-based models with labels }
    \label{fig:resultc}
\end{subfigure}
\caption{Illustration of Macro F1-scores of models for various K-shot settings. Vertical bars denote standard deviation of results over 5 seeds. HateBERT is ommited to simplify comparison. }
\label{fig:all_results}
\end{figure*}

Macro F-1 score is chosen as the evaluation metric. We first discuss the performance of models relative to each other in their respective setting, then provide a top-down analysis. Figure \ref{fig:all_results} illustrates the performance  for each setting. Detailed numerical results  are displayed in Table \ref{apdx:all_results} of the Appendix. We also provide Figure \ref{fig:all_results_apdx} as an alternative illustration to facilitate comparison between models.

\begin{table*}
\small
\begin{center}
\begin{tabular}{lcccccccc}
\toprule
\textbf{} & \multicolumn{2}{c}{\textbf{ToxiGen}} & \multicolumn{2}{c}{\textbf{HateXplain}} & \multicolumn{2}{c}{\textbf{Implicit\_hate}} & \multicolumn{2}{c}{\textbf{Covid}} \\
{} & \textbf{F1} & \textbf{$\sigma$} &    \textbf{F1} & \textbf{$\sigma$} &   \textbf{F1} & \textbf{$\sigma$} & \textbf{F1} & \textbf{$\sigma$} \\
\midrule
\textbf{HateBERT} & 0.731   & 0.007   &  0.571   & 0.007  &  0.523        & 0.010  & 0.542   &  0.020 \\ \midrule
\textbf{RoBERTa\_untrained    } &   0.753 &  0.010 &      0.592 &  0.011 &         \textbf{0.604} &  0.018 &   0.571 &  0.014 \\
\textbf{RoBERTa\_binary       } &   0.754 &  0.015 &      \textit{0.621} &  0.014 &         0.578 &  0.011 &   0.559 &  0.022 \\
\textbf{RoBERTa\_retrained    } &   \textbf{0.795} &  0.004 &      0.527 &  0.043 &         0.592 &  0.013 &   0.596 &  0.014 \\ \midrule
\textbf{ProtoNet             } &   0.721 &  0.009 &      0.520 &  0.003 &         0.387 &  0.005 &   0.452 &  0.018 \\
\textbf{ProtoMAML            } &   0.677 &  0.105 &      0.553 &  0.041 &         0.552 &  0.021 &   0.452 &  0.159 \\
\textbf{MLDG                 } &   0.750 &  0.011 &      0.569 &  0.028 &         0.579 &  0.006 &   0.568 &  0.027 \\ \midrule
\textbf{ProtoNet\_Token       } &   0.691 &  0.021 &      0.541 &  0.018 &         0.380 &  0.023 &   0.469 &  0.015 \\
\textbf{ProtoNet\_Label       } &   0.516 &  0.048 &      0.338 &  0.033 &         0.146 &  0.011 &   0.228 &  0.012 \\
\textbf{ProtoNet\_Full        } &   0.693 &  0.006 &      0.529 &  0.008 &         0.274 &  0.017 &   0.406 &  0.016 \\ \midrule
\textbf{JE\_ProtoNet          } &   0.751 &  0.010 &      0.610 &  0.015 &         0.569 &  0.007 &   0.615 &  0.008 \\
\textbf{JE\_ProtoNet\_Untrained} &   \textit{0.758} &  0.008 &      0.575 &  0.011 &         \textit{0.595} &  0.014 &   \textit{0.617} &  0.011 \\
\textbf{JE\_ProtoNet\_CLS      } &   \textit{0.758} &  0.005 &      \textbf{0.628} &  0.007 &         \textit{0.595} &  0.018 &   \textbf{0.636} &  0.031 \\
\bottomrule
\end{tabular}
\end{center}
\caption{Macro F1-scores and their standard deviation ($\sigma$) for K = 256. Highest and second-highest F1-scores in each test domain are \textbf{bolded} and \textit{italicized}, respectively. }
\label{tab:k256}
\end{table*}

\begin{table*}[!ht]
\begin{tabular}{l|c|c|c|cc|cc|cc|}
\cline{2-10}
 &
  \multirow{2}{*}{\textbf{Train size}} &
  \multirow{2}{*}{\textbf{No. class}} &
  \multirow{2}{*}{\textbf{F1 Max}} &
  \multicolumn{2}{c|}{\textbf{64}} &
  \multicolumn{2}{c|}{\textbf{128}} &
  \multicolumn{2}{c|}{\textbf{256}} \\ \cline{5-10} 
                                 &        &   &       & \multicolumn{1}{c|}{F1\%} & Size \% & \multicolumn{1}{c|}{F1\%} & Size\% & \multicolumn{1}{c|}{F1\%} & Size\% \\ \hline
\multicolumn{1}{|l|}{\textbf{ToxiGen}}    & 512\textsuperscript{*}   & 2 & 0.795 & \multicolumn{1}{c|}{91}   & 25      & \multicolumn{1}{c|}{93}   & 50     & \multicolumn{1}{c|}{95}   & 100    \\ \hline
\multicolumn{1}{|l|}{\textbf{HateXplain}} & 16,118 & 3 & 0.687 & \multicolumn{1}{c|}{79}   & 1       & \multicolumn{1}{c|}{84}   & 2      & \multicolumn{1}{c|}{91}   & 5      \\ \hline
\multicolumn{1}{|l|}{\textbf{Implicit\_hate}} &
  3,807 &
  6 &
  0.586 &
  \multicolumn{1}{c|}{88} &
  10 &
  \multicolumn{1}{c|}{96} &
  20 &
  \multicolumn{1}{c|}{102} &
  39 \\ \hline
\multicolumn{1}{|l|}{\textbf{Covid}}      & 16,000 & 4 & 0.832 & \multicolumn{1}{c|}{53}   & 2       & \multicolumn{1}{c|}{66}   & 3      & \multicolumn{1}{c|}{76}   & 6      \\ \hline
\end{tabular}
\caption{Comparison of JE\_ProtoNet\_CLS performance and size across K values \{64, 128, 256\}. \textit{F1 \%} represents the model's F1-score relative to the highest F1-score (\textit{F1 max}) reported by the original authors using the corresponding training data size (\textit{Train size}). \textit{Size \%} indicates the sample size percentage based on the K-value relative to the \textit{Train size}.  \textsuperscript{*}Best F1-score attained by \textit{RoBERTa\_binary} at K=256 chosen (statistics not reported by original authors) } 
\label{tab:size}
\end{table*}

\subsection{For Baseline Settings}
Results for Baseline experiments are displayed in Figure \ref{fig:resulta}. Unsurprisingly, macro F1-scores improve with less variance as the number of K training shots increases. With the exception of ToxiGen's binary classification, models tend not to attain most of their classifying capability until K=128. \textit{RoBERTA\_untrained}, which uses no pre-training, displays a  consistently improvement in performance with more data across all 4 test domains. In contrast, pre-training on in-domain data with the Mask Language Model objective, causes underperformance in some domains (HateXplain, Implicit\_hate), while provides a boost for others at various K's (ToxiGen, Covid). Pre-training on binary collapsed data allows \textit{Roberta\_binary} to attain better F-1 scores in low-resource cases (K < 64), suggesting beneficial initialization from exposure to general data on offensive speech. Nevertheless, this method does not guarantee peak performance when more in-domain training data is available. This finding aligns with prior cross-domain hate speech experiments \citep{fortuna2021well, toraman2022large}, suggesting that the binary mapping scheme might overlook specific nuances unique to each domain, hindering generalization to new domains. Overall, these simple pre-training methods offer inconsistent performance.

\subsection{For Meta-Learning Settings}
\subsubsection{Without Label}
\label{sec:without_label}
Optimization-based models require more training data (K$\ge64$) to exhibit competitive performance. \textit{Proto\_MAML}'s F1-scores are inferior to those of \textit{MLDG} in every setting. Furthermore, this model displays considerably more dispersed results between seeds than the others, especially at higher K values for ToxiGen and Covid. These factors suggest that relying on the inductive bias using prototype-based initialization of the classifier may not enhance generalization between domains. On the other hand, \textit{MLDG}, specifically designed for domain generalization, appears to perform comparably to \textit{Robeta\_binary} at the extremes of K values, with similar trajectory in between. Nevertheless, this model's performance also displays higher standard deviation for K $\in$ \{128, 256\} for HateXplain and Covid. 

\textit{ProtoNet} has the  distinction of offering the best F1-scores at K=16 for all test domains and consistently stable results for different seeds, thanks to its metric-based nature. However, this feature also appears to hamper its classifying power even when exposed to more in-domain training data,  showing little improvement at maximum value of K. 

\subsubsection{With Label}
\label{sec:with_label}
Though all \textit{ProtoNet}-based models exhibit relatively stable performance across evaluation seeds as in previous setting, their trajectories differ when trained on more in-domain data (Figure \ref{fig:resultc}). Interestingly, \textit{ProtoNET\_Label}'s performance deteriorates as K increases, along with higher standard deviation in comparison to other variants.

Appending the entire definition to the input also does not appear to be viable, as  \textit{ProtoNet\_Full} yields the second least favorable F-1 scores for HateXPlain, Implicit\_hate, and Covid domains. \textit{ProtoNet\_token}, though yielding more favorable results compared to the 2 previous variants, do not demonstrate significant difference in performance compared to the base \textit{ProtoNet} setting in \ref{sec:without_label}.

\textit{JE\_ProtoNet} is the only model whose performance appreciatively scales with increment of K values. This architecture achieves competitive F1-scores at K $\in $ \{16, 32\}, with notably higher result for Implicit\_hate. More importantly, \textit{JE\_ProtoNet}  outperforms other with-label variants across all test domains, demonstrating its  robustness.

\noindent \textbf{Does pre-training help initialize Joint Embedding?} We perform testing on the \textit{JE\_ProtoNet} model, whose Attention module's weights are randomly initialized, without any pre-training on the 10 datasets, denoted as \textit{JE\_ProtoNet\_Untrained}. In Figure \ref{fig:resultc}, we observe that \textit{JE\_ProtoNet\_Untrained}'s F1-scores are inferior to that of its counterpart \textit{JE\_ProtoNet} across domains for K$\le$128, except for \textit{Implicit\_hate} domain at K=64. Additionally, the former generally exhibits higher variance among results compared to the latter model for K$\le$64. These notions suggest that our pre-training approach via meta-learning provides advantageous initialization when in-domain resource is scarce. 

\noindent \textbf{Leveraging JE\_ProtoNet's features} Observing the discrepancy in improvement with more resources (higher K-shots) of Baseline models, and ProtoNet-based models' good performance in low-resource setting, we hypothesize that it is possible to enhance \textit{JE\_ProtoNet} to overcome its limited inductive bias while fully utilizing its learned discriminate features. We thus equip \textit{JE\_ProtoNet} with a classification head, a feed forward neural network that takes the [CLS] token form the joint embedding as input. This model, \textbf{JE\_ProtoNet\_CLS}, is discussed in the next section.

\subsubsection{Global Assessment} 
We restrict our analysis here to K=256. From Table \ref{tab:k256}, we observe that all models perform respectably on \textit{ToxiGen}'s  classification task. This finding is in line with the dataset's conditional machine-generation of its binary labels. \textit{HateBERT}'s performance generally trails behind our baseline models, indicating this model's struggle to adapt to new domains in few-shot settings. Interestingly, pre-training on in-domain data allows RoBERTa to leading F1-score of 0.795. On the other hand, \textit{RoBERTa\_untrained} achieves the leading score of 0.604 on \textit{Implicit\_hate}. Nevertheless, none of the baseline models obtain consistently good performance across the board. Meta-learning models without labels also do not produce competitive results. This group's top performer, \textit{MLDG}, attains only decent results across test domains.

As discussed in \ref{sec:with_label}, \textit{ProtoNet} with various methods to incorporate label information do not yield improvement over their non-label counterpart, and may even exhibit degradation (\textit{ProtoNet\_Full}). Using joint embedding that incorporates label definitions, however, achieves both strong and consistent F-1 scores, as shown by all configurations of \textit{JE\_ProtoNet} models. In fact, \textit{JE\_ProtoNet\_CLS} attains best or second best results in all 4 test domains, especially the 0.636  F1-score for \textit{Covid}, arguably the most semantically distinctive domain.

\section{Discussion}
\noindent \textbf{Definition matters} While many works in offensive speech literature have focused on standard classification techniques, ours is the first to leverage the definition of associated labels. Our proposed framework to incorporate definition via the joint embedding is beneficial to boost classification performance over other models, given the same amount of training data. In addition to enhancing annotation quality, this factor is yet another signal to encourage researchers to pay more attention to their terminologies to both enhance downstream tasks and facilitate cross-task studies. 

\noindent \textbf{More data is not always needed } Our experiments provide a case study on how much data is necessary to achieve certain results in the area of offensive speech detection. While having more labeled data is always preferable, the annotation process can be expensive, and thus constituting a barrier to researchers not equipped with abundant resources. Table \ref{tab:k256} describes the percentages of \textit{JE\_ProtoNet\_CLS}'s F1-scores for K=64 to 256 relative to the F1-scores reported by the original authors using the entire training data. Most notably, our balanced data sampling and model design achieve 79\% of max F1-score with 1\% training data for \textit{HateXplain}, 76\% with 6\% training data for \textit{Covid}, and even bested the max F1-score with only 39\% training data for \textit{Implicit\_Hate}. 

\noindent \textbf{Recommendations for Low-Resource Settings} This observation suggests that tailoring the data annotation process for class balance may allow offensive speech classifiers to attain better performance with less training resource. For instance, practitioners may opt to iterate over collecting, annotating, testing and increasing the quantity of data using classification metrics as guiding criteria. Experimental results suggest that setting K = 64 may be a good starting point. As our technique does not incur significant technical overhead compared to baseline architectures, researchers may implement both to mutually juxtapose during this iterative data collection process, and stop when the models' performances plateau or reach a satisfactory threshold. This approach has the advantage of being both data-efficient and empirically driven.

\section{Conclusion}
While we also leverage the existing rich corpora, our work explores a different setting of offensive speech detection compared to other works, such as HateBERT or fBERT \citep{caselli2021hatebert, sarkar2021fbert}. 
The proposed joint-embedding may be adapted to complement other existing architectures. Our approach can also be applied to other NLP tasks, such as sentiment analysis and stance detection, where labels extend beyond compact phrases. We invite researchers to explore definitions and the extent of their usefulness in other tasks.

Offensive content is ever-evolving in today's world. We hope that our findings provide useful pointers for NLP practitioners to more efficiently explore diverse topics in this field. 

\section*{Limitations}
Our pre-processing step that removes special characters and casts inputs into lower-case is chosen for efficiency and to facilitate fair comparisons between the various experimental configurations. It is possible that these characters provide additional predictive signal, and could be used to enhance the models' performance. 

This work uses RoBERTa as the sole backbone architecture for our models. In recent years, a plethora of new, potentially more powerful architectures have been proposed and may obtain better performance on our tasks. Furthermore, our corpora all focus on English, which does not reflect the diversity of languages, cultural norms and expressions that can express offensive sentiment. Our classification tasks only explore label categories, while other works also explicitly predict the targets of offensive content. Finally, definition for label is not always available for all offensive speech datasets. It remains an open research question if our method will transfer to other domains, not limited to offensive speech.  We invite interested researchers to explore these venues. 

\section*{Ethics Statement}
This research aims to reduce the spread of offensive content by means of more reliably detecting them. Our compiled datasets do not violate privacy as they are extracted from published works, whose authors have taken steps to uphold confidentiality. We acknowledge that, due to the open nature of this data, they might contain references to real life personnel. There exists a risk that nefarious parties may leverage the ideas proposed in this work in the opposite of the authors' intention to propagate more offensive speech instead.


\bibliography{anthology,custom}
\bibliographystyle{acl_natbib}

\appendix

\section{Appendix}
\label{apdx}

\begin{table*}[!ht]
\small
\centering
\begin{tabular}{|l|p{1cm}|l|p{2cm}|p{4.5cm}|}
\hline
Model & Meta Epoch & Meta Learning Rates & Finetune Epoch & Finetune Learning Rates \\
\hline
Baselines & - & - & 3 & 2e-5 \\ \hline
MLDG & 5 & E:5e-5; C:1e-4 & 3 & E:2e-5 ; C:\{-1:5e-3, 16:5e-3, 32:5e-3, 64:7e-3, 128:7e-3, 256:5e-4\} \\ \hline
ProtoMAML & 5 & E:5e-5; C:1e-4 & 4 & E:2e-5 ; C:\{-1:1e-3, 16:1e-4, 32:1e-4\} \\ \hline
ProtoNet (all variants) & 5 & E:2e-5; C:1e-4 & 2 & E:1e-5 \\ \hline
JE\_ProtoNet & 5 & E:5e-5; A:2e-5; C:1e-4 & 3 & E:2e-5, A:2e-5 \\ \hline
JE\_ProtoNet\_CLS & - & - & 3 & E:2e-5, A:2e-5; C:1e-4 \\ 
\hline
\end{tabular}
\caption{Hyperparameter values chosen for reported runs. For learning rates, E stands for \textit{Word Embedding}, (RoBERTa), A for \textit{Attention} module, C for \textit{Classification} head. If no letter specififed, then learning rate applies to all components. Learning rates in bracketed dictionaries are tied to the corresponding component, with the key represents the corresponding K-shot value it is applied to. -1 denotes the default rate.}
\label{apx:hyper}
\end{table*}

\section*{Preprocessing}
\label{apx:preprocess}
We perform standard preprocessing steps on our data. First, we remove non-ASCII characters from the inputs and convert them to lower case. Special platform-specific characters are removed, with certain exceptions (e.g. hyperlinks replaced with \textit{<url>}, user-mentions with \textit{<user>}, hashtags are segmented into separate tokens by using the Ekphrasis Python library \footnote{Available at https://github.com/cbaziotis/ekphrasis}). We also replace repetitive patterns with a single representative (e.g. \textit{“b b b ”} to \textit{“b”}).

\section*{Technological Details}
\label{apx:tech}
All models are trained using single NVIDIA P100 GPU, with the exception on JE models, which were trained on NVIDIA A100 GPU. Our system also posses 20GB of RAM memory.

To select hyperparameters for the K-shot fine-tuning process on test domains, we use a sample of size K=64 from the left over data after the initial K-shot training samples. To select hyperparameters during meta-training, we monitor the average losses and F1-score during meta-testing. Meta-learning models are trained over a number of meta epochs, where each consists of 300 tasks randomly chosen from the 10 training domains. For fine-tuning, the learning rate is equipped with the Cosine Annealing Learning Rate scheduler \footnote{As implemented by Pytorch library} with minimum rate set to 1e-5. \textit{Roberta\_retrained} models are pre-trained using MLM objective for 5 epochs. Learning rates are chosen from the following pool of candidates: \{1e-5, 2e-5, 5e-5, 7e-5, 1e-4, 5e-4, 7e-4, 1e-3\}. Fine tuning and meta epochs are chosen from \{2,3,4,5\}. Batch sizes are set to 16. Table \ref{apx:hyper} shows the final values of hyperparameters.

\begin{table*}[!htp]
\small
\begin{center}
\begin{tabular}{p{2.5cm}p{12cm}}
\toprule  Dataset &  Definition \\
\midrule \citealp{waseem2016hateful} & \textbf{Overtly Aggressive} : any text in which aggression is overtly expressed either through the use of specific kind of lexical items or lexical features which is considered aggressive and or certain syntactic structures is overt aggression. \textbf{Covertly Aggressive} : any speech in which aggression is overtly expressed either through the use of specific kind of lexical items or lexical features which is considered aggressive and or certain syntactic structures is overt aggression. \textbf{Non Aggressive} : any text that does not fall into the other two categories. \\ \hline

\citealp{golbeck2017large} &  \textbf{Offensive} : uses a sexist or racial slur, attacks a minority, seeks to silence a minority, criticizes a minority without a well founded argument , promotes, but does not directly use, hate speech or violent crime, criticizes a minority and uses a straw man argument, blatantly misrepresents truth or seeks to distort views on a minority with unfounded claims, shows support of problematic hash tags, negatively stereotypes a minority, defends xenophobia or sexism, contains a screen name that is offensive, as per the previous criteria, the tweet is ambiguous at best , and the tweet is on a topic that satisfies any of the above criteria. \textbf{Not Offensive} : does not into any other categories. \\ \hline

\citealp{davidson2017automated} &  \textbf{Targeted Insult} : posts containing insult threat to an individual, a group, or others. \textbf{Untargeted Insult} : posts containing non targeted profanity and swearing. posts with general profanity are not targeted, but they contain non acceptable language. \textbf{Not Offensive} : posts that do not contain offense or profanity \\ \hline

\citealp{kumar2018aggression} & \textbf{Harrassment} : deeply racist, misogynistic or homophobic, or otherwise bigoted. the use of shocking language primarily to upset the person who is reading. unapologetically or intentionally offensive this could be someone saying something with the intent of upsetting a group, or an extreme account e.g. neo nazis using language that they approve of but they know the general public would disapprove of. have language intended to make the target or a broader group fearful or to feel unsafe. express hate or extreme bias to a particular group. could be based on religion, race, gender, sexual orientation. language directed at a particular person or group designed to upset them. this language may be milder than in other cases but should be part of the campaign by one person or a group to make the target feel threatened or intimidated. \textbf{Not Harrassment} : anything that does not rise to the level of clearly and unambiguously fitting into the other categories. \\ \hline

\citealp{founta2018large}  &  \textbf{Hate Speech} : targeting immigrants; content must have immigrants refugees as main target, or even a single individual, but considered for his her membership in that category and not for the individual characteristics ; must deal with a message that spreads, incites, promotes or justifies hatred or violence against target, or a message that aims at dehumanizing, hurting or intimidating the target. or expresses hating towards women in particular in the form of insulting, sexual harassment, threats of violence, stereotype, objectification and negation of male responsibility. \textbf{Not Hate Speech} : the followings are not considered hate speech, against other target, offensive language, blasphemy, historical denial, over incitement to terrorism, offense towards public servant, defamation. \\ 
\hline
\citealp{zampieri2019predicting} &  \textbf{Abusive Language} : any strongly impolite, rude or hurtful language using profanity, that can show a debasement of someone or something, or show intense emotion. \textbf{Hate Speech} : language used to express hatred towards a targeted individual or group, or is intended to be derogatory, to humiliate, or to insult the members of the group, on the basis of attributes such as race, religion, ethnic origin, sexual orientation, disability, or gender. \textbf{Normal} : tweets that do not fall in any of the other categories \\ \hline

\citealp{basile2019semeval} &  \textbf{Hate Speech} : language that is used to expresses hatred towards a targeted group or is intended to be derogatory, to humiliate,or to insult the members of the group. may also be language that threatens or incites violence. \textbf{Offensive Language} : may contain offensive terms but targets disadvantaged social groups in a manner that is potentially harmful to them. \textbf{Neither} : language that does not all into either of the other categories .\\ \hline

\citealp{sap2019social} &  \textbf{Offensive} : denotes the overall rudeness, disrespect, or toxicity of a post. whether a post could be considered offensive to anyone.\textbf{ Not Offensive} : not offensive to anyone. \\ \hline

\citealp{toraman2022large}  &  \textbf{Hate} : target, incite violence against, threaten, or call for physical damage for an individual or a group of people because of some identifying trait or characteristic. \textbf{ Offensive} : humiliate, taunt, discriminate, or insult an individual or a group of people in any form, including textual. \textbf{Normal} : does not fall into any of the other categories .\\ \hline

\citealp{vidgen2021learning} &  \textbf{Derogation} : content which explicitly attacks, demonizes, demeans or insults a group. \textbf{Animosity} : content which expresses abuse against a group in an implicit or subtle manner. \textit{Threatening} : content which expresses intention to, support for, or encourages inflicting harm on a group, or identified members of the group. \textbf{Support For Hateful Entities} : content which explicitly glorifies, justifies or supports hateful actions, events, organizations, tropes and individuals collectively, entities. \textbf{Dehumanization} : content which perceives or treats people as less than human. \textbf{Not Hate} : content that falls into none of the other categories .\\
\bottomrule
\end{tabular}
\caption{Full definition for train datasets}
\label{apx:train_def}
\end{center}
\end{table*}

\begin{table*}[!ht]
\small
\begin{center}
\begin{tabular}{p{2.5cm}p{12cm}}
\toprule  Dataset &  Definition \\
\midrule  Covid &  \textbf{Hostility Against An East Asian Entity} : express abuse or intense negativity against an east asian entity, primarily by derogating attacking them. also includes conspiracy theories, claiming east asians are a threat, and expressing negative emotions about them. \textbf{Criticism Of An East Asian Entity} : make a negative judgement assessment of an east asian entity, without being abusive. this includes commenting on perceived social, economic and political faults, including questioning their response to the pandemic and how they are governed. \textbf{Discussion Of East Asian Prejudice} : tweets that discuss prejudice related to east asians but do not engage in, or counter, that prejudice. includes content which discusses whether east asian prejudice has increased during covid 19, the supposed media focus on prejudice, and or free speech. also includes explicitly challenge or condemn abuse against an east asian entity. includes rejecting the premise of abuse, describing content as hateful or prejudicial, or expressing solidarity with target entities. \textbf{None Of The Above} : do not fall into any of the other categories. could be abusive in other ways, such as expressing misogyny. \\ \hline
ToxiGen &  \textbf{Toxic} : text that could be seen as harmful to anyone, may contain offensive rude, humor, insults, personal attacks, profanity, aggression, may refer to targeted group with harmful intent that is expressed in stereotypes or lewd manners. minority groups include blacks, asians, native americans, muslims, latinos, jewish, chinese, lgbtq , mentally disabled, physically disabled, mexican, women and middle eastern. \textbf{Benign} : text that could not be seen as harmful or intended to be harmful to anyone. \\ \hline
Implicit\_hate & \textbf{White Grievance} : includes frustration over a minority group s perceived privilege and casting majority groups as the real victims of racism. this language is linked to extremist behavior and support for violence. \textbf{Incitement To Violence} : includes flaunting in group unity and power or elevating known hate groups and ideologies. \textbf{Inferiority Language }: implies one group or individual is inferior to another, and it can include dehumanization denial of a person s humanity , and toxification language that compares the target with disease, insects, animals . related to assaults on human dignity, dominance, and declarations of superiority of the in group. \textbf{Irony} : refers to the use of sarcasm , humor, and satire to attack or demean a protected class or individual. \textbf{Stereotypes And Misinformation} : associate a protected class with negative attributes such as crime, or terrorism. includes misinformation that feeds stereotypes and vice versa, like holocaust denial and other forms of historical negationism. \textbf{Threatening And Intimidation }: conveys a speaker's commitment to a target s pain, injury, damage, loss, or violation of rights, threats related to implicit violation of rights and freedoms, removal of opportunities, and more subtle forms of intimidation. \\ \hline
HateXplain &  \textbf{Hate Speech} : language which attacks, demeans, offends, threatens, or insults a group based on race, ethnic origin, religion, disability, gender, age, sexual orientation, or other traits. it is not the presence of certain words that makes the text hate speech, rather you should look the context the word is used in the text. \textbf{Offensive Language} : usage of rude, hurtful, derogatory, obscene or insulting language to upset or embarasse people. \textbf{Normal} : neither hate speech nor offensive .\\
\bottomrule
\end{tabular}
\end{center}
\caption{Full definition for test datasets}
\label{apx:test_def}
\end{table*}

\begin{figure*}
  \centering
  \begin{subfigure}{\textwidth}
    \centering
    \includegraphics[width=\linewidth]{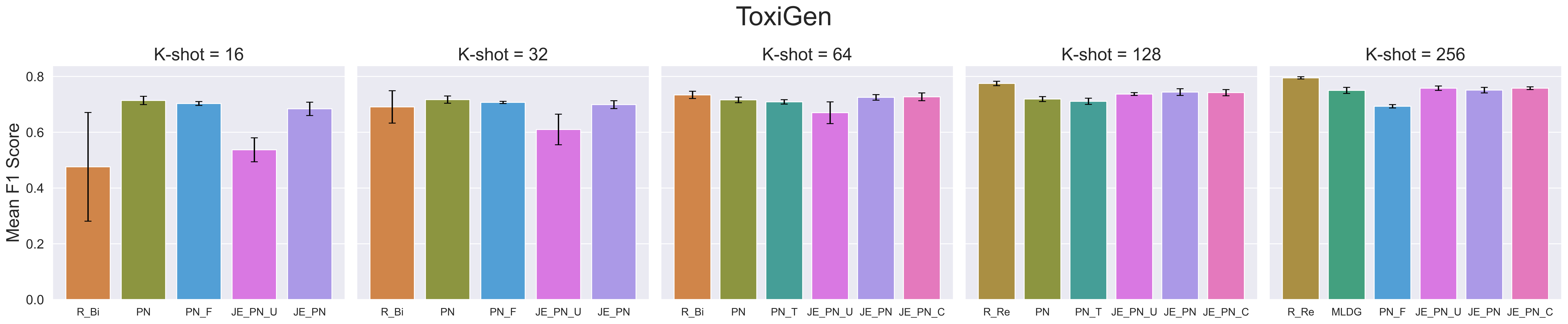}
    \label{fig:sub1}
  \end{subfigure}%
  \hfill
  \begin{subfigure}{\textwidth}
    \centering
    \includegraphics[width=\linewidth]{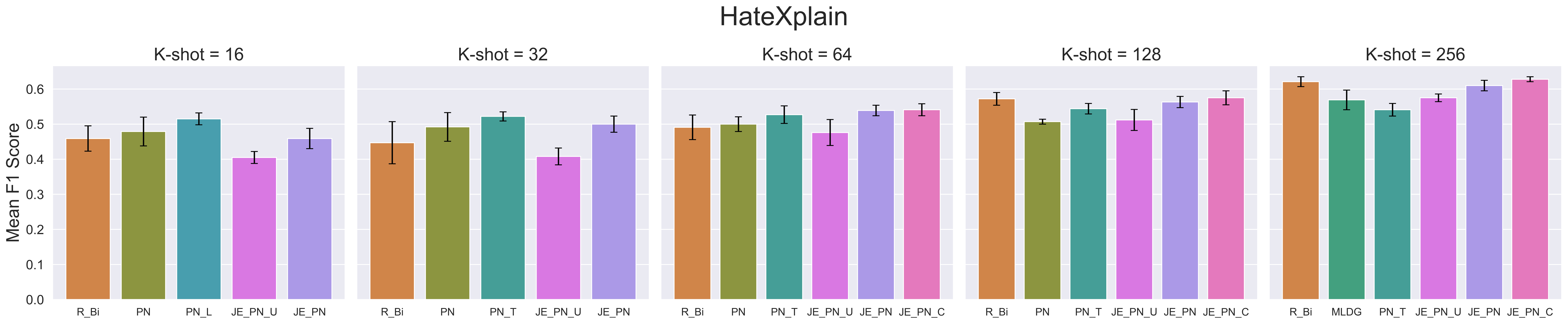}
    \label{fig:sub1}
  \end{subfigure}%
  \hfill
  \begin{subfigure}{\textwidth}
    \centering
    \includegraphics[width=\linewidth]{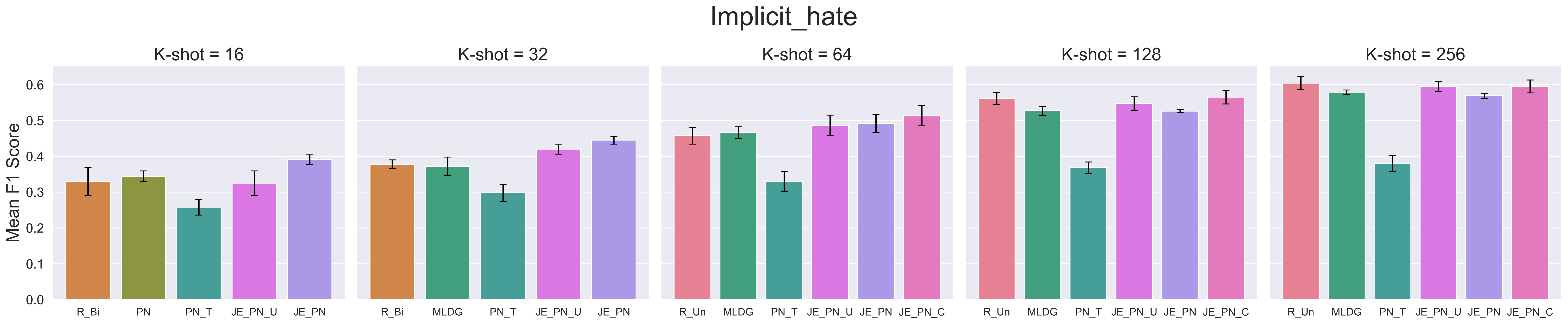}
    \label{fig:sub1}
  \end{subfigure}%
  \hfill
    \begin{subfigure}{\textwidth}
    \centering
    \includegraphics[width=\linewidth]{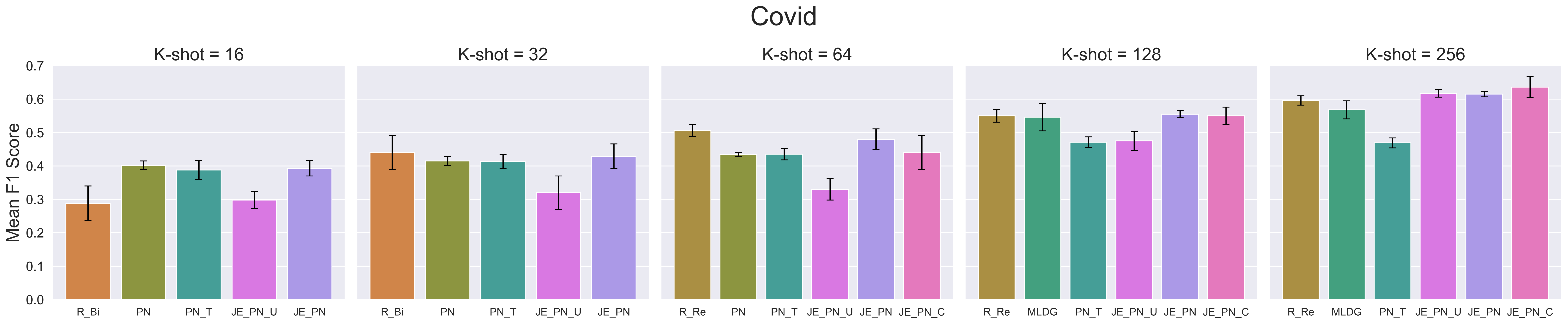}
    \label{fig:sub1}
  \end{subfigure}%
  \hfill
  \caption{Mean macro F-1 scores of various models on 4 test sets at different K-shot settings, with error bars representing standard deviation over 5 seeds . For each K, the first bar shows the best performer among the \textit{Baseline} models, the second bar shows the best among the models \textit{Without Label}, and the third among models \textit{With Label}. The rest includes all applicable \textit{Joint-Embedding} models. \textit{R\_Bi}: RoBERTa\_binary, \textit{R\_Re}: RoBERTa\_retrained, \textit{R\_Un}: RoBERTa\_untrained, \textit{PN}: ProtoNet,  \textit{PN\_F}: ProtoNET\_Full, \textit{PN\_T}: ProtoNet\_Token, \textit{PN\_L}: ProtoNet\_Label, \textit{PM}: ProtoMAML, \textit{JE\_PN}: JE\_ProtoNet, \textit{JE\_PN\_U}: JE\_ProtoNet\_Untrained, \textit{JE\_PN\_C}: JE\_ProtoNet\_CLS.}
  \label{fig:all_results_apdx}
\end{figure*}

\begin{table*}
\small
\begin{center}
\begin{tabular}{llcccccccc}
\toprule
{} & {} & \multicolumn{2}{c}{\textbf{ToxiGen}} & \multicolumn{2}{c}{\textbf{HateXplain}} & \multicolumn{2}{c}{\textbf{Implicit\_hate}} & \multicolumn{2}{c}{\textbf{Covid}} \\
{\textbf{K}} & {\textbf{Model}} &  \textbf{F1} & \textbf{$\sigma$} &    \textbf{F1} & \textbf{$\sigma$} &   \textbf{F1} & \textbf{$\sigma$} & \textbf{F1} & \textbf{$\sigma$} \\

\midrule
\multirow{12}{*}{\textbf{16 }}     & \textbf{HateBERT} & 0.462  & 0.095  & 0.326      & 0.032  &  0.233        & 0.047   & 0.368   & 0.037  \\
    & \textbf{RoBERTa\_untrained} &   0.377 &  0.075 &      0.210 &  0.090 &         0.232 &  0.052 &   0.184 &  0.108 \\
    & \textbf{RoBERTa\_binary} &   0.476 &  0.195 &      0.459 &  0.036 &         0.330 &  0.039 &   0.288 &  0.052 \\
    & \textbf{RoBERTa\_retrained} &   0.427 &  0.144 &      0.171 &  0.018 &         0.181 &  0.010 &   0.171 &  0.127 \\
    & \textbf{ProtoNet} &   \textbf{0.714} &  0.015 &      0.479 &  0.041 &         \textit{0.344} &  0.015 &   \textbf{0.402} &  0.013 \\
    & \textbf{ProtoMAML} &   0.343 &  0.020 &      0.180 &  0.016 &         0.086 &  0.025 &   0.201 &  0.002 \\
    & \textbf{MLDG} &   0.557 &  0.112 &      0.293 &  0.053 &         0.263 &  0.049 &   0.314 &  0.040 \\
    & \textbf{ProtoNet\_Token} &   \textit{0.695} &  0.019 &      \textit{0.502} &  0.022 &         0.258 &  0.022 &   0.388 &  0.028 \\
    & \textbf{ProtoNet\_Label} &   0.668 &  0.049 &      \textbf{0.515} &  0.017 &         0.221 &  0.021 &   0.364 &  0.014 \\
    & \textbf{ProtoNet\_Full} &   0.703 &  0.007 &      0.489 &  0.027 &         0.246 &  0.020 &   0.357 &  0.038 \\
    & \textbf{JE\_ProtoNet} &   0.684 &  0.024 &      0.459 &  0.029 &         \textbf{0.391} &  0.013 &   \textit{0.393} &  0.023 \\
    & \textbf{JE\_ProtoNet\_Untrained} &   0.537 &  0.043 &      0.405 &  0.017 &         0.325 &  0.034 &   0.298 &  0.025 \\
    & \textbf{JE\_ProtoNet\_CLS} &      -- &     -- &         -- &     -- &            -- &     -- &      -- &     -- \\
\cline{1-10}
\multirow{12}{*}{\textbf{32 }} & \textbf{HateBERT} & 0.498  & 0.092  & 0.373      & 0.044  &  0.359        & 0.010   & \textit{0.429}   & 0.028  \\
    & \textbf{RoBERTa\_untrained} &   0.482 &  0.073 &      0.365 &  0.078 &         0.318 &  0.043 &   0.323 &  0.053 \\
    & \textbf{RoBERTa\_binary} &   0.691 &  0.058 &      0.447 &  0.060 &        0.378 &  0.012 &   \textbf{0.440} &  0.051 \\
    & \textbf{RoBERTa\_retrained} &   0.574 &  0.134 &      0.192 &  0.069 &         0.231 &  0.049 &   0.273 &  0.138 \\
    & \textbf{ProtoNet} &   0.717 &  0.013 &      0.492 &  0.041 &         0.369 &  0.023 &   0.415 &  0.014 \\
    & \textbf{ProtoMAML} &   0.368 &  0.074 &      0.191 &  0.026 &         0.177 &  0.077 &   0.205 &  0.014 \\
    & \textbf{MLDG} &   0.637 &  0.044 &      0.384 &  0.050 &         0.372 &  0.026 &   0.367 &  0.038 \\
    & \textbf{ProtoNet\_Token} &   \textit{0.706} &  0.009 &      \textbf{0.522} &  0.013 &         0.298 &  0.024 &   0.413 &  0.021 \\
    & \textbf{ProtoNet\_Label} &   0.694 &  0.016 &      \textit{0.517} &  0.006 &         0.222 &  0.014 &   0.338 &  0.018 \\
    & \textbf{ProtoNet\_Full} &   \textbf{0.707} &  0.004 &      0.498 &  0.028 &         0.251 &  0.019 &   0.382 &  0.011 \\
    & \textbf{JE\_ProtoNet} &   0.699 &  0.014 &      0.500 &  0.023 &         \textbf{0.445} &  0.011 &   \textit{0.429} &  0.037 \\
    & \textbf{JE\_ProtoNet\_Untrained} &   0.610 &  0.055 &      0.408 &  0.024 &         \textit{0.420} &  0.014 &   0.320 &  0.050 \\
    & \textbf{JE\_ProtoNet\_CLS} &      -- &     -- &         -- &     -- &            -- &     -- &      -- &     -- \\
\cline{1-10}
\multirow{12}{*}{\textbf{64 }} & \textbf{HateBERT} & 0.558  & 0.109  & 0.466      & 0.017  &  0.420        & 0.011   & 0.455   & 0.031  \\
& \textbf{RoBERTa\_untrained} &   0.498 &  0.148 &      0.422 &  0.084 &         0.457 &  0.023 &   0.430 &  0.061 \\
    & \textbf{RoBERTa\_binary} &   \textbf{0.734} &  0.013 &      0.491 &  0.035 &         0.424 &  0.032 &   \textit{0.494} &  0.021 \\
    & \textbf{RoBERTa\_retrained} &   0.688 &  0.088 &      0.177 &  0.026 &         0.342 &  0.047 &   \textbf{0.506} &  0.018 \\
    & \textbf{ProtoNet} &   0.716 &  0.010 &      0.500 &  0.021 &         0.375 &  0.017 &   0.434 &  0.006 \\
    & \textbf{ProtoMAML} &   0.442 &  0.156 &      0.318 &  0.078 &         0.263 &  0.083 &   0.202 &  0.019 \\
    & \textbf{MLDG} &   0.680 &  0.018 &      0.393 &  0.063 &         0.467 &  0.017 &   0.385 &  0.041 \\
    & \textbf{ProtoNet\_Token} &   0.709 &  0.008 &      0.527 &  0.025 &         0.329 &  0.028 &   0.435 &  0.017 \\
    & \textbf{ProtoNet\_Label} &   0.669 &  0.045 &      0.500 &  0.049 &         0.196 &  0.019 &   0.310 &  0.027 \\
    & \textbf{ProtoNet\_Full} &   0.702 &  0.004 &      0.507 &  0.029 &         0.266 &  0.019 &   0.390 &  0.008 \\
    & \textbf{JE\_ProtoNet} &   \textit{0.725} &  0.010 &      \textit{0.539} &  0.015 &         \textit{0.491} &  0.025 &   0.480 &  0.031 \\
    & \textbf{JE\_ProtoNet\_Untrained} &   0.670 &  0.039 &      0.476 &  0.037 &         0.486 &  0.029 &   0.330 &  0.032 \\
    & \textbf{JE\_ProtoNet\_CLS} &   0.727 &  0.014 &      \textbf{0.541} &  0.017 &         \textbf{0.513} &  0.028 &   0.441 &  0.051 \\
\cline{1-10}
\multirow{12}{*}{\textbf{128}} & \textbf{HateBERT} & 0.687  & 0.012  & 0.537      & 0.009  &  0.497        & 0.010   & 0.490   & 0.027  \\
    & \textbf{RoBERTa\_untrained} &   0.707 &  0.032 &      0.528 &  0.029 &         \textit{0.561} &  0.017 &   0.534 &  0.026 \\
    & \textbf{RoBERTa\_binary} &   0.744 &  0.020 &      \textbf{0.572} &  0.018 &         0.535 &  0.013 &   0.539 &  0.019 \\
    & \textbf{RoBERTa\_retrained} &   \textbf{0.775} &  0.008 &      0.332 &  0.073 &         0.515 &  0.030 &   \textit{0.550} &  0.019 \\
    & \textbf{ProtoNet} &   0.719 &  0.009 &      0.507 &  0.007 &         0.379 &  0.011 &   0.445 &  0.018 \\
    & \textbf{ProtoMAML} &   0.551 &  0.143 &      0.400 &  0.107 &         0.448 &  0.022 &   0.340 &  0.078 \\
    & \textbf{MLDG} &   0.712 &  0.013 &      0.473 &  0.041 &         0.527 &  0.013 &   0.546 &  0.041 \\
    & \textbf{ProtoNet\_Token} &   0.711 &  0.011 &      0.544 &  0.015 &         0.368 &  0.016 &   0.471 &  0.016 \\
    & \textbf{ProtoNet\_Label} &   0.576 &  0.041 &      0.420 &  0.014 &         0.164 &  0.016 &   0.280 &  0.033 \\
    & \textbf{ProtoNet\_Full} &   0.702 &  0.010 &      0.519 &  0.016 &         0.247 &  0.014 &   0.410 &  0.015 \\
    & \textbf{JE\_ProtoNet} &   \textit{0.744} &  0.012 &      \textit{0.563} &  0.016 &         0.526 &  0.004 &   \textbf{0.555} &  0.010 \\
    & \textbf{JE\_ProtoNet\_Untrained} &   0.737 &  0.005 &      0.512 &  0.030 &         0.547 &  0.019 &   0.475 &  0.029 \\
    & \textbf{JE\_ProtoNet\_CLS} &   0.742 &  0.011 &      0.575 &  0.020 &         \textbf{0.565} &  0.019 &   \textit{0.550} &  0.026 \\
\bottomrule
\end{tabular}
\end{center}
\caption{Macro F1-scores of models on 4 test domains with K=16 to 128. Best performance for each K per dataset in \textbf{bold}, second best \textit{italicized}.}
\label{apdx:all_results}
\end{table*}

\end{document}